\documentclass[conference]{IEEEtran}
\usepackage{array}
\newcolumntype{P}[1]{>{\centering\arraybackslash}p{#1}}
\IEEEoverridecommandlockouts
\usepackage{amsmath,amssymb,amsfonts}
\usepackage{algorithmic}
\usepackage{graphicx}
\usepackage[para,online,flushleft]{threeparttable}
\usepackage{textcomp}
\usepackage{xcolor}
\usepackage{biblatex}
\def\BibTeX{{\rm B\kern-.05em{\sc i\kern-.025em b}\kern-.08em
    T\kern-.1667em\lower.7ex\hbox{E}\kern-.125emX}}
\begin{document}

\title{Leveraging Transfer Learning for Reliable Intelligence Identification on Vietnamese SNSs (ReINTEL) }

\author{\IEEEauthorblockN{Trung-Hieu Tran}
\IEEEauthorblockA{\textit{Zalo Group - VNG Corporation} \\
Ho Chi Minh City, Vietnam \\
hieutt9@vng.com.vn}
\\
\IEEEauthorblockN{Truong-Son Nguyen}
\IEEEauthorblockA{
\textit{Zalo Group - VNG Corporation} \\
\textit{University of Science, VNU-HCMC} \\
Ho Chi Minh City, Vietnam \\
ntson@fit.hcmus.edu.vn}
\and
\IEEEauthorblockN{Long Phan}
\IEEEauthorblockA{\textit{Zalo Group - VNG Corporation} \\
Ho Chi Minh City, Vietnam}

\\
\IEEEauthorblockN{Tien-Huy Nguyen}
\IEEEauthorblockA{
\textit{Zalo Group - VNG Corporation} \\
\textit{University of Science, VNU-HCMC} \\
Ho Chi Minh City, Vietnam \\
huynt@vng.com.vn }
}

\maketitle

\begin{abstract}
This paper proposed several transformer-based approaches for  Reliable Intelligence Identification on Vietnamese social network sites at VLSP 2020 evaluation campaign. We exploit both of monolingual and multilingual pre-trained models. Besides, we utilize the ensemble method to improve the robustness of different approaches. Our team achieved a score of 0.9378 at ROC-AUC metric in the private test set which is competitive to other participants.

\end{abstract}


\section{Introduction}
With the emerging of social networks sites in the last decade, the information shared on social network sites (SNSs) gaining attention significantly.
SNSs provide platforms for users to share their own content, react, or add comments on the content posted by other users. They help strangers to be connected based on their common interests, activities, identities, or professions \cite{info:doi/10.2196/jmir.8382}. However, issues arise as these information tend to be spread in a rapid and free manners. Many contemporary mainstream news outlets has reported and been cautious that these information can be unreliable, and even political institutions around the world have brought up an emphasis to curb the phenomenon \cite{scott_eddy_2017}. These false information can create a dreadful misunderstand of any particular worldwide event. Therefore, the need for detecting unreliable SNSs through any site or platform has gain unprecedented attention in order to prevent any spreading of these misleading information \cite{DBLP:journals/corr/RuchanskySL17,DBLP:journals/corr/abs-1712-07709}. In this paper we offer a contribution in leveraging multiple pre-trained transformer models to validate identifying shared information in Vietnamese SNSs. 

\section{Methodology}
There are several pre-trained Transformer-based models in Vietnamese language. They divide into two categories: multilingual such as BERT Multilingual \cite{devlin2018bert}, XLM-RoBERTa, \cite{conneau2020unsupervised} and monolingual such as PhoBERT \cite{phobert}, vELECTRA \cite{the2020improving}, vBERT \cite{the2020improving}. In this section, we present our approach to fine-tune the above Transformer-based models for reliable intelligence identification task. Our source code\footnote{https://github.com/heraclex12/VLSP2020-Fake-News-Detection} is publicly available.

\subsection{Data Pre-processing}
We process the text contents in two phases. First, we tokenize the contents using TweetTokenizer from the NLTK toolkit\footnote{https://www.nltk.org/} and use the emoji\footnote{https://pypi.org/project/emoji/} package to translate icons into text strings. Word segmentation is required in some pre-trained models, we use VnCoreNLP \cite{vu-etal-2018-vncorenlp} to tokenize the input. In the text content, several features can affect model performance. To clarify this, we duplicate the texts into two versions, one is lower-case texts and no newline characters, the other is raw texts.

In the second phase, to fine-tune the Transformer-based models such as BERT Multilingual, viBERT, and viELECTRA, we must insert [CLS] special token and [SEP] special token to the input. The [CLS] is encoded including all representative information of the whole input sentence. Meanwhile, the use of [SEP] token is simply to separate different sentences of an input. In our method, we only need to insert the [SEP] token to the end of every input. Finally, we convert the new input to the sequence of indices with the same length. We will pad the sequences with the [PAD] token if their length is less than the specified length. In PhoBERT and XLM-RoBERTa, instead of [CLS], [SEP], and [PAD], we use $<$s$>$, $<$/s$>$, $<$pad$>$ respectively.

\subsection{Models}
\subsubsection{Single Models}\label{singmodel}
We leverage the ability of pre-trained Transformer-based models, which are trained on large-scale datasets by a variety of unsupervised learning methods. In this task, we list some popular models in the Vietnamese language. They divide into two categories following:
\begin{itemize}
    \item Multilingual: BERT Multilingual employs masked language model and next sentence prediction for pre-training. Meanwhile, XLM-RoBERTa relies on masked language modeling objective and cross-lingual language modeling objective without next sentence pre-training objective. Both models trained on a large-scale dataset with multiple languages.
    \item Monolingual: viBERT uses the same architecture as BERT Multilingual. However, it just trained on Vietnamese dataset. viELECTRA is released with viBERT. viELECTRA uses a new pre-training task, called replaced token detection. PhoBERT is also masked language model, which trained on 20GB Vietnamese corpus.
\end{itemize}

We examine the diversity of input lengths for each of the above models including 256-length, 512-length, and multiple 512-length as a long document. Handling long document is inspired by the research in \cite{9003958}, we segment the input into multiple chunks with the length of 512 and feed them into the Transformer-based model to obtain contextual representations. Then we propagate each output through a Bi-LSTM layer to get document embedding. Finally, we perform the final classification in a linear layer.

\begin{figure}[!t]
    \centering
    \includegraphics[width=0.5\textwidth,height=8.5cm]{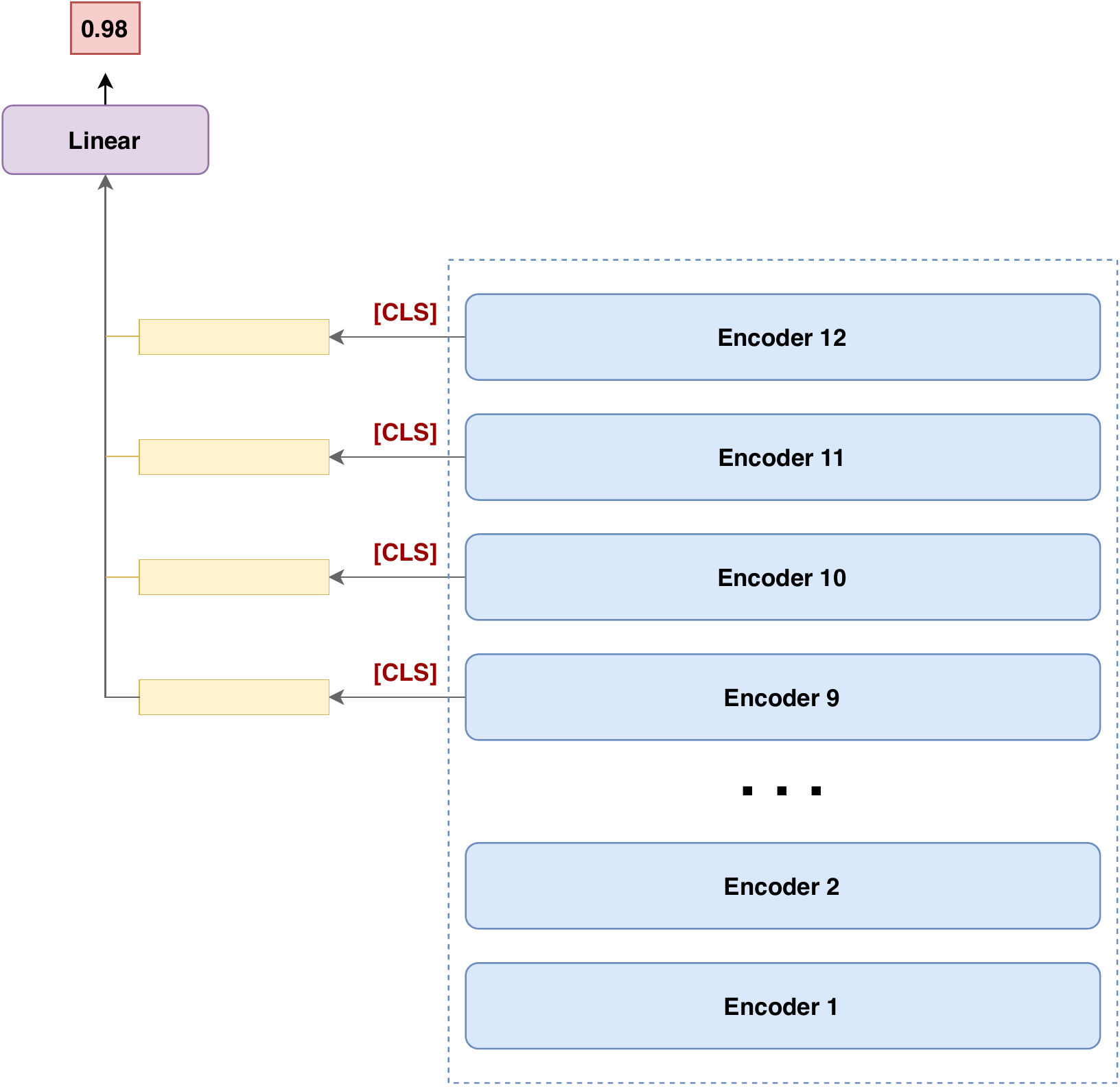}
    \caption{Proposed Neural Network Architecture}
    \label{fig:model}

\end{figure}

We stack a linear classifier on the top of output representations. Instead of just using the last hidden layer, based on the result in \cite{devlin2018bert}, we concatenate the last four hidden layers as the input of linear classifier, this modification might slightly improve model performance. For more clearly, Figure \ref{fig:model} illustrates the architecture of our model in detail.

\subsubsection{Ensemble Method}
To utilize the robustness of the different Transformer-based models such as viBERT, viELECTRA, PhoBERT. We select three different Transformer-based models with the highest ROC-AUC score in the validation set. Then, we average their label probabilities to get final probabilities.

\section{Experiments}
\subsection{Dataset}
The dataset given the VLSP 2020's Organizer\footnote{https://vlsp.org.vn/vlsp2020/eval} contains 4,372 training examples and about 1,642 examples for each public test set and private test set. Each example includes some information such as the encoded id of the owner, the text content of the post, the number of likes, shares, comments, and some photos. This dataset is really imbalanced. The unreliable class accounts for about 17\% of the training set, the reliable class dominates with about 83\%. This dataset also contains nearly 10\% long texts with more than 500 tokens.

\subsection{Experimental Setup}
For training, we use Adam optimizer with a learning rate of 3e-5 and a weight decay of 0.01. Initially we freeze all layers of Transformer-based model to make sure the gradients are not calculated in the first epoch. We initialize each of the models described in Section \ref{singmodel} with 10 different random states, then we train them for 20 epochs and select the models with the highest ROC-AUC score in the validation set for predicting.

\subsection{Results}
In public test set, all our models trained on lower-case texts. Some models fail in the validation set. Thus, we do not use them for predicting. As can be seen from Table \ref{table:1}, the monolingual models such as viBERT, viELECTRA, and PhoBERT achieve better performance than the multilingual models. The effectiveness of length is also different for each model architecture. Specifically, viBERT with the length of 512 is worse than 256-length. Meanwhile, viELECTRA with the length of 512 significantly outperforms 256-length viELECTRA, about 3.21\% improvement. 512-length viELECTRA also overcomes all single models, it achieves the highest score in both public test and private test. 

In order to leverage the ability of the ensemble method, we select three best models with different styles including 256-length viBERT, 512-length viELECTRA, and 256-length PhoBERT to average their output probabilities, we refer this as 3-Ensemble model. In addition, we also denote the top six models with 6-Ensemble. This experiment result demonstrates the ensemble method can further boost the performance.

\def\arraystretch{1.5}%
\setlength{\tabcolsep}{0.72em} %
\begin{table}[htbp]
\begin{threeparttable}
\centering
\caption{The results on the public test set}
\begin{tabular}{ p{4cm}|P{1.6cm}|P{1.6cm}   }
 \hline
    \textbf{Model} & \textbf{Length} & \textbf{ROC-AUC} \\
\hline
    $BERT Multilingual$ & 512 & 83.36 \\
    & multiple 512 & 88.67\\
    $viBERT$ & 256 & 91.44 \\
    & 512 & 90.20\\
    & multiple 512 & 89.35 \\
    $viELECTRA$ & 256 & 89.37\\
    & 512 & 92.58 \\
    $PhoBERT$ & 256 & 91.79 \\
    $3-Ensemble_{base}$& & \textbf{92.80} \\
    $6-Ensemble_{base}$ & & 92.20 \\

 \hline
\end{tabular}
\label{table:1}
\end{threeparttable}
\end{table}

\def\arraystretch{1.5}%
\setlength{\tabcolsep}{0.72em} %
\begin{table}[htbp]
 \begin{threeparttable}
\centering
\caption{The results on the private test set}
\begin{tabular}{ p{4cm}|P{1.6cm}|P{1.6cm}  }
 \hline
    \textbf{Model} & \textbf{Length} & \textbf{ROC-AUC} \\
\hline
    $3-Ensemble_{base,uncased}$ & & 91.90 \\
    $viELECTRA_{cased}$ & 512 & 93.09 \\
    $viELECTRA_{feature,cased}$ & 512 & \textbf{93.78} \\
    $viELECTRA_{uncased}$ & 512 & 91.60 \\
    $PhoBERT_{cased}$ & 256 & 93.01\\
    $PhoBERT_{uncased}$ & 256 & 91.99\\

 \hline
\end{tabular}
\label{table:2}
\end{threeparttable}
\end{table}

After reviewing the data, arbitrary capitalization makes the text content unprofessional. In private test set, we investigate the influence of letter case on model performance. viELECTRA with the length of 512 and PhoBERT with the length of 256 achieve the highest score in the public test set, so we decide to train them on the raw texts which remain the upper-case and newline character. The results are described in Table \ref{table:2} shows that the cased models significantly improve the uncased models. The ROC-AUC values being 1.49\% and 1.02\% above the uncased models for viELECTRA and PhoBERT. If only based on text content, it is difficult to identify fake news. In common, the news sources will be an important factor to distinguish reliable information. Because of that, we deliver the username feature into our models. The usernames with high unreliable examples will be penalized, which reduces their output probabilities by a certain value and vise versa. This intuition brings a significant improvement to 512-length cased viELECTRA, from 93.09\% to 93.78\%.

\section{Conclusion and Future work}
We conduct extensive experiments to investigate the robustness of multiple Vietnamese pre-trained Transformer-based models to solve the Reliable Intelligence Identification on Vietnamese SNSs problem. Experimental results show that transfer learning method can provide high efficiency in detecting fake news. Along with that is the potential of the ensemble method can be exploited.

Because the nature of news itself can be hard to predict even for humans, in future work, we propose a plan to integrate an external resource checking to validate the source of the news and evaluate this as weights in the models. The news with validated or authorize resources like governments or accredited sites will have higher weights as a piece of real news and vice versa. Besides, we also intend to take advantage of features in images which can provide much evidence for identifying reliable information.

\printbibliography
\end{document}